\documentclass[12pt]{iopart}

\begin{document}
\title{First steps to a constructor theory of cognition}
\author{Riccardo Franco}
\address{riccardofrancoq@gmail.com}
\date{\today}
\begin{abstract}
This article applies the conceptual framework of constructor theory of information to cognition theory. The main result of this work is that cognition theory, in specific situations concerning for example the conjunction fallacy heuristic, requires the use of superinformation media, just as quantum theory. This result entails that quantum and cognition theories can be considered as elements of a general class of superinformation-based subsidiary theories. 
\end{abstract}
 \maketitle
\section{Introduction}
Quantum-like cognition is a quite recent research field making use of concepts and methods taken from quantum theory and quantum probability to describe a variety of judgements and decision making findings which are considered \textit{irrational} \cite{gilovich2002heuristics}. Such approach follows the intuition that human reasoning - in a wide range of situations relevant to bounded-rationality - can be described more conveniently by using the laws of quantum rather than classical probability. Quantum-like cognition seems to take into account puzzling effects like conjunction fallacies \cite{franco2009conjunction}, disjunction fallacies, averaging effects, unpacking effects, and order effects \cite{busemeyer2011quantum}, violations of sure-thing principle in decision theory \cite{pothos2009quantum},  violations of symmetry in similarity judgements \cite{pothos2011quantum} and to produce new predictions like the quantum question (QQ) equality \cite{wang2014context}. Despite its successful applications, in last years some experimental data evidenced that the theory needs a deeper understanding. In particular, the violation of the Grand-reciprocity (GR) equations \cite{boyer2016testing} and the problems relevant to question order effects and response replicability \cite{khrennikov2014quantum}. An even deeper critique arrives from \cite{pleskac2013s}, where it is noted that the same formalism has been used (for example in conjunction fallacy experiments) to describe both judgements (subjective probabilities) and choice tasks frequencies  (objective probabilities). The authors thus argue that \textit{to suppose these two models are of the same class seems to be an error}. Moreover, 

It is clear that we are at the first steps of the theory, and many points need to be analysed carefully, like for example the internal structure of vector spaces representing a concept and the mathematical tools able to distinguish between knowing an event or giving an answer about it. A first attempt in such direction has been made in \cite{franco2016newtheory}. But the main question is \textit{why and how should we use quantum formalism}?  Actual quantum-like cognition models clearly evidence the quantum-like nature of the model: they simply use the mathematical formalism of quantum formalism to describe situations concerning cognitive psychology. This approach is agnostic for what concerns \textit{the physical nature of the mental processes}. We stress the fact that at the moment there aren't certain data which evidence the quantum nature of physical processes in human mind. However, it is important to clearly consider the question if, in the description of cognitive psychological processes, \textit{the quantum formalism is really necessary}. From a phenomenological point of view, actual results in quantum-like cognition suggest that the quantum-like models allow to describe quite naturally some cognitive effects, but is this the only possible way? From one side, the computational complexity involved in cognitive processes suggests that using only classical resources to perform  a task like playing chess would require an unreasonable amount of time. The features of quantum parallelism and the resulting quantum speedup evidenced by quantum algorithms suggest that somehow in human cognitive processes the quantum formalism could play an important role. 

However, these intuitions are not conclusive, and stronger arguments are needed. They should come from constructor theory of information. An important result of such theory is that, if perfect cloning operation is not possible in general, the class of subsidiary theories we need is one using  superinformation media, and quantum theory is an instance of it. Constructor theory of information is used in tyhe present article to embed cognitive psychological main definitions and concepts into general definitions of the theory. This allows to put in evidence the role of cognitive process and the importance of the cloning task, which helps to discriminate between two \textit{cognitive regimes}: one based on classic information, and one based on superinformation.  

This article is organized as follows.  We first introduce the constructor theory as a fundamental theoretical layer allowing to present cognition as a subsidiary theory. In section \ref{sec_representation} we introduce the cognitive representation of concepts in terms of attributes and variables, two basic elements in constructor theory. We put such elements in correspondence with definitions already existing in the field of cognitive science (features). In section \ref{sec_tasks} we discuss another basic element of constructor theory, the task, in terms of cognitive process. 
A precise distinction between subjects' answers and the knowledge of uncertain  events is provided in section \ref{sec_judgements}, evidencing an ambiguity of previous quantum-like cognition models, as suggested in \cite{franco2016newtheory}. Such distinction leads to treat in different ways subjective probabilities (judgement) and answer frequencies. In particular, the new model clearly distinguishes between subjective and objective probabilities, focusing on the concept of subject's knowledge. Finally, after translating into cognitive terms the basic tasks defined in constructor theory of information \cite{deutsch2015constructor}, we show  in section \ref{sec_conj_super} that there are specific situations where cloning tasks are not possible, thus deriving the necessity to use a kind of information, the superinformation, which is directly connected with the quantum formalism. In other words, we show that a theory based on superinformation is necessary for a wide range of cognitive tasks. This result puts all cognitive theory in a new and different perspective, allowing for the introduction of subsidiary theories which are in the same class of quantum theory.

\section{Constructor theory for cognition}
Constructor theory is a general conceptual tool which is able to describe other theories, also called \textit{subsidiary theories}. For example, constructor theory has been applied to information, leading thus to \textit{constructor theory of information}. This theory reconciles two apparently contradictory features of information: from one side of being an abstraction (yet governed by laws of physics), from the other of being physical (yet counter-factual). Moreover, it robustly unifies the theories of quantum and classical information. Similarly, the study of cognitive processes from one side concerns with physical transformations, but from the other it is relevant to information.  For these reasons, constructor theory is a useful theory in the context of cognition, helping to identify classical and quantum-like features of human behaviour. 

\textit{Constructor theory} (or more precisely construction-task theory) \cite{deutsch2013constructor} describes the world in terms of transformations involving two kinds of physical systems, playing different roles. The first is the \textit{constructor}, which is the agent causing the transformation, remaining unchanged in its ability to cause new transformations. The second consists of the subsystems -which we refer to as the \textit{substrates}- which are transformed from having some physical attribute to having another. 
The basic principle of constructor theory of information \cite{deutsch2015constructor} is that \textit{all other laws of physics are expressible entirely in terms of statements about which physical transformations are possible and which are impossible, and why}. So a task is \textit{possible} in this sense when it could be performed with arbitrarily high accuracy (not that it will happen with non-zero probability). Unitary quantum theory is a class of superinformation theories. 
A fundamental result of the theory is that superinformation theories exhibit unpredictability as a consequence of the impossibility of cloning certain sets of attributes. On the contrary, the class of \textit{information theories} are defined by allowing the perfect cloning of information. Classic information theory is an instance of information theories.

It becomes clear the importance of the constructor theory in the context of cognitive psychology: if we re able to identify situations where the cloning task is not possible, we can conclude that cognition theory requires the use of superinformation. This would entail the necessity to develop a theory of cognition based on superinformation, just like quantum theory: a quantum-like theory of cognition thus is a good candidate.  Constructor theory provides a unifying approach for other scientific theories: its principles constrain their laws, and in particular, require certain types of task to be possible. In this perspective, we introduce the basic definitions and principles of constructor theory, and then we apply them in the context of cognition, which can be viewed as a \textit{subsidiary theory}. The fundamental principle of constructor theory of cognition is the following: \textit{all (other) laws of human cognition are expressible solely in terms of statements about which cognitive processes are possible, which are impossible, and why}. This will be cleared in the following. Here we underline the fundamental role of physics in the description of cognition:  any cognitive representation is a specific configuration of one or more physical systems, which make part of subject's brain.  Moreover, any cognitive process is always a physical transformation, which evidences the importance of studying  how such cognitive can be performed.

%
\section{Constructors: subjects}
The \textit{constructor} is the agent causing the transformation, whose defining characteristic is that it remains unchanged in its ability to cause the transformation again. No perfect constructors exist in nature. Approximations to them, such as catalysts or robots, have non-zero error rates and also deteriorate with repeated use. 

In cognitive context, the constructor is a subject who thinks.  A subject for example can simply think, or he can make a choice or give an answer: all these situations cause a transition of belief. In other situations, the interaction is with an external system, also called \textit{environment}, like for example another subject or an object. The subject is able to perform again cognitive processes leaving unchanged such ability, of course with human limits imposed by attention and physical resources.

It is important to note that constructor theory does not focus on constructors, but on the possibility to construct tasks. 

\section{Substrates: cognitive resources}
In constructor-theoretic physics the primitive notion of a physical system is replaced by the slightly different notion of  \textit{substrate} — a physical system some of whose properties can be changed by a physical transformation. It is clear that the focus of such definition is the physical transformations: in other words, the importance of a physical system consists in its ability to be changed by some transformation.

Substrates will be denoted as capital boldface  letters. 

In cognition theory, the substrate is represented by the physical systems in the brain and also in other parts of human body whose properties (determininig the subjects thoughts) can be changed by a physical transformation. It is known, in fact, that subjects thoughts are influenced by information stored in the brain or coming from senses, but also from emotions, and their relation is strong.

It is important to stress that this approach is very general and it does not make initial hypotheses about the form or organization if such substrate. We only identify as a substrate a physical system in the system \textit{brain plus body} that can be changed by a physical transformation and thus the place where cognitive processes occur.

The initial hypothesis is that \textit{for any cognitive process it is possible to identify a specific substrate (or set of substrates) where a physical process (corresponding to that cognitive process) occurs.}


In physics it is easy to identify a substrate and to distinguish a substrate from another. For example, we can distinguish a particle from another, or an object from another. In cognition, instead, it is difficult to identify different physical systems into the brain in terms of corrwsponding physical processes. Recent results \cite{huth2016natural} evidence, using the semantic space as a visualization tool, that categories can be represented in incredibly intricate maps that cover much more of the brain than expected. In other words, when a subject listens to a word, many different areas of the brain seem to be activated.
This is consistent with the hypothesis that collective modes in the brain are excited when a cognitive process occurs.

However, as we will see further, human reasoning requires the ability to manage copies of systems representing an information. Moreover, subjects need new cognitive resources every time they learn something new. This fact leads us to consider the potential existence of many different substrates in the brain.

In summary, we need to keep a general description of substrates in cognition theory, avoiding additional hypotheses and allowing the existence of several substrates. It is clearly possible to define aggregations of substrates, which are again substrates.
\section{States, attributes and variables}\label{sec_representation}
Any subsidiary theory must provide a collection of \textit{states}, \textit{attributes} and \textit{variables}, for any given substrate. In the present case, we note that cognition is the theory describing subjects' beliefs and knowledge. We can say that in cognition theory the state of a substrate is the \textit{subject's mental state} (or belief). In constructor theoretic terms, thinking to something means assigning specific properties to a substrate, the subject's belief: thus the \textit{states} of a substrate are the possible configurations of properties of such substrate.  We will write states with small Greek letters, like for example $\psi$, and when necessary we will add the subscript which refers to the specific substrate, like for example $\psi_{\textbf{A}}$.  

We now introduce the basic concept of  \textit{attribute}, which identifies all the states in which a specific property is true. Starting from the context of physics, we give two examples, one taken from classic information, the other from quantum theory.
\begin{itemize}
\item A traffic light: each of the three lamps (red, amber, green) can be in a state $\sigma_{i}$ (with $i=r, a, g$). Each lamp  can present only the states \textit{on} or \textit{off}: there aren't any other possible situations. Thus these states can be defined in terms of the attributes $1,0$ which describe the activation of the lamp. In general, a traffic light is a substrate whose eight states are labelled by a binary string $(\sigma_r,\sigma_a,\sigma_g)$. 
\item  In quantum physics, two attributes of an electron are \textit{the values $\pm 1/2$ for the $z$-component of the spin}: there are electron states for which this component is $+1/2$, other states for which is $-1/2$, but there are also states for which such attributes are not definite. In fact superpositions or mixtures of these two states represent a different kind of states, where the considered property in not always true, but only possible. 
\end{itemize}
In cognitive context, the attribute is a property given by all subjective mental states where that property is considered true with certainty. Thus we can say that \textit{an attribute identifies all the belief states for which such attribue is a known property}. Attributes are also called \textit{features}, whose meaning in cognitive psychology is described in the following.  We now provide some examples.
\begin{itemize}
\item The knowledge of the property \textit{on} of a specific lamp of a traffic light defines the attribute \textit{lamp certainly on}. The subject's belief state of a traffic light after looking at the traffic light describes the knowledge of the lamp states. This can be one of the eight states  labelled by a binary string $(\sigma_r,\sigma_a,\sigma_g)$. However, if the subject can't look at the traffic light, his knowledge is uncertain and the belief state is in none of these eight states.
\item The knowledge of the outcome of a coin toss \textit{head} or \textit{tail} is an attribute. The subject's belief state of a tossed coin can thus be \textit{knowledge of head} or \textit{ knowledge of tail}, which define the correspoding attributes, also labelled as 0 and 1. But if the subject does not know the result to the flip, his state is uncertain and can't be identified by attributes 0 or 1.
\end{itemize} 

A \textit{variable} is defined as a set of attributes of the same substrate, like for example \textit{red, spherical, with legs}, which provides a specific meaning. In cognitive theory, variables represent sets of certain knowledge states. It is easy to recognize that such variables are  \textit{concepts, categories, events}: in other words, sets of attributes which help to represent and carachterize such concepts. The structure and nature of such collections is the focus of \textit{categorization}. Before briefly describing the main approaches to categorization, we provide some further remarks and examples. In cognitive context we can define also variables describing not external objects, but elements completely relevant to the subject. For example, the knowledge of subject's answer (yes or no) to a question is a variable whose value is provided by the subject himself; similarly, the knowledge of subject's judged probabilities form variables, whose attributes are the value ranges of the probabilities. 
We finally say that a variable $X$ is \textit{sharp}  with the value $x$ on a specific substrate state when such state has attribute $x \in X$, thus when there is certain knowledge about a specific attribute in the set defining $X$. 
\begin{itemize}
\item The collection of possible subject's knowledge states about the   outcomes of a coin toss represents a variable. If the subject knows with certainty the result of the flip (head for instance), the variable on such belief state is sharp.
\item A card in a standard card deck has variables \textit{seed} and \textit{value}: knowing the seed and the value, we can identify univocally a card. In this case, there are no constraints between the two variables.
\item Considering a digital camera, the producer and the sensor dimension are two possible variables. It is clear that the knowledge of a specific producer may constrain the sensor dimension. 
\item The most part of categories describing real word elements present different variables with different membership degrees.  The  concept of \textit{chair} presents some core variables, such as \textit{we can seat on it}  (which can be considered sharp), and other, like \textit{with four legs}, \textit{made in wood} (which are not sharp). As we will see in the following, the category \textit{chair} is a prototypical concept, whose structure is described by prototype theory.
\item The title of the famous book of George Lakoff \textit{Women, fire and other dangerous things} describes a category found in  Dyirbal (australian aboriginal) language. As we can see, the set of attributes identifying  a category may change from a country to another, from an hystorical period to another, from a set of subjects to another. 
\end{itemize}

It is clear that the description of an object can be more or less detailed. What in a first description are attributes may become new variables, because the previous attributes is now considered as a collection of other attributes. In the example of the card deck, the numbers can be grouped into even/odd values, which may be considered as two variables.

\subsection{Categories in constructor theory}
Even if in different ways,  the most important approaches to categorization of concepts make use of attributes. They are called \textit{features}, and they are used by subjects to represent some property of an object, an event, a concept. For this reason, we will use attributes and features as synonims. Following \cite{murphy1985role}, we call such approaches feature-based theories, since features are the key ingredient to compute the similarities between two concepts. In the following, we briefly recall two basic strategies to approach the representation problem \cite{murphy1985role}: the classical view and the probabilistic view. 
The first and simplest representation strategy is the \textit{classical} approach, where concepts are defined by singly necessary and jointly sufficient features.  The classical Aristotelian view claims that categories are discrete entities characterized by a set of properties which are shared by their members: these properties are assumed to establish the conditions which are both necessary and sufficient conditions to capture meaning, establishing the basis for natural taxonomy. In different contexts, such approach is used for organize information, leading to the concept of  \textit{enumerative} scheme \cite{broughton2004faceted, rosenfeld2002information}, which divide the information into ever smaller classes according to identified principles of division. This may be considered a top-down approach, as the ambition is to partition the overall corpus into narrower segments until the content of each segment consistently describes the same concept. The alphabetical organization of the phone book’s white pages is a simple example.

The second approach defines a common core of criterial properties and argues that concepts may be represented in terms of features that are typical or characteristic, rather than defining.  Such approach is also called the \textit{probabilistic} view or \textit{feature-based} view. 
According to Rosch's \textit{prototype theory}, concepts are organized around family resemblances, and consist of characteristic, rather than defining features. These features are weighted in the definition of the prototype. Rosch showed that subjects rate conceptual membership as graded, with degree of membership of an instance corresponding to conceptual distance from the prototype. Moreover, the prototype appears to be particularly resistant to forgetting. Prototype theory also has the strength that it can be mathematically formulated and empirically tested. A measure of the \textit{conceptual distance} between the instance and the prototype can be obtained by calculating the similarity between the prototype of a concept and a possible instance of it, across all salient features, or by considering exemplars. 
Such approach has some similaities with the \textit{analytic-synthetic} (also called \textit{faceted}) strategy \cite{ranganathan1985faceted}, which seeks to identify the constituent concepts for each element and from the resultant set of concepts evolve schemes which arrange these concepts within a classificatory structure. Faceted classification is also called \textit{multidimensional} classification, since it allows to find items based on more than one dimension, and can be considered a bottom-up approach.  In fact, the faceted classification describes an object/event in terms of different features (facets), which can be considered as the relevant dimensions. The mutually exclusive instances of a facet are called \textit{foci}. Usually, there are constraints on how the foci can be combined into the compound concepts. 
For example, some subjects shopping for jewelry may be most interested in browsing by particular type of jewelry (earrings, necklaces), while others are more interested in browsing by a particular material (gold, silver). Material and type are examples of facets; earrings, necklaces, gold, silver are examples of facet values, but if necessary new facets could be added to allow a simpler browsing. Another classic example are the factes for wines: the color (red, white, rosè), the region, the producer, the price range, dry/sweet. The value red for the color of course acts as a constraint on the facet dry/sweet. It is known that about 69\% of websites makes at least some use of faceted classification in their search interfaces, which reveals the importance of such approach and how much it is natural for subjects. 

It is important to observe that, even in the probabilitic view, an attribute-matching perspective is not able to capture a complete perspective. In \cite{murphy1985role} it is noted that probabilistic view relies directly or indirectly on the notion of similarity, and thus it fails in various ways to represent intra- and inter-concept relations and more general world knowledge. First, if similarity is the sole explanation of category structure, then an immediate problem is that the similarity relations among a set of entities depend heavily on the particular weights given to individual features. A barber pole and a zebra would be more similar than a horse and a zebra if the feature \textit{striped} had sufficient weight. The point is that any two entities can be arbitrarily similar or dissimilar by changing the criterion of what counts as a relevant attribute. A second major critique is that, given two objects, we can find an infinite list of features in common and also an infinite list of differences. For example, plums and lawnmowers are very different, but both weigh less than 10,000 kg (and less than 10,001 kg, . . .), both did not exist 10,000,000 years ago (and 10,000,001 years ago,. . .), both cannot hear well, both can be dropped, both take up space, and so on. The approach suggested in \cite{murphy1985role}, also called \textit{theory theory},  is focused on people's theories about the world: in other words, the concepts are organized by theories, where we use the term "theory" to mean any of a host of mental "explanations," rather than a complete, organized, scientific account. It is easy to show examples showing that simple family resemblance is not able to capture the typicality structure of goal-derived categories. For instance, let us consider the category that includes the objects children, jewelry, portable TVs, paintings, manuscripts, and photograph albums. These objects have low family resemblance. However, once the theme taking things out of one's home during a fire is known, these judgments become easy. Notice that this concept is not a "natural" one, yet it does seem to hang together in its context. Such examples suggest that theories can elucidate the relations among very different objects and thereby form them into a coherent category, even if they do not form a "natural" class.


\section{Tasks: cognitive processes}\label{sec_tasks}
Constructor theory's primitive elements are \textit{tasks} (as defined below), which intuitively can be thought of as the specifications of physical transformations affecting substrates. Its laws take the form of conditions on possible/impossible tasks on substrates allowed by subsidiary theories. In cognitive context, clearly tasks are \textit{cognitive processes} (or cognitive tasks) that are responsible of the transformations of these beliefs: in other words, they represent human thoughts.

A cognitive task is the abstract specification of a cognitive transformation on a belief (substrate), which is transformed from having some cognitive attribute to having another. It is expressed as a set of ordered pairs of input/output attributes $x_i\rightarrow y_i$ of the substrates
\begin{equation}
\{x_1\rightarrow y_1, x_2 \rightarrow y_2, .. \}
\end{equation}
The substrates with the input attribute are presented to the constructor, which delivers the substrates with the output attribute. A constructor is capable of performing a task if, whenever presented with the substrates with a legitimate input attribute of the task (i.e. in any state in that attribute), it delivers them in some state in one of the corresponding output attributes, regardless of how it acts on the substrate with any other attribute. 

Let us now consider some basic examples.
\begin{itemize}
\item If $x_1, x_2$ are the sides of a coin (\textit{head}, \textit{tail}), the transition $x_1 \rightarrow x_1$, $x_2 \rightarrow x_1$ represents a cognitive process which changes an undefined state in a sharp state ($x_1$) relevant to the variable $X$. This may describe the knowledge of the result of a coin toss, or the simple act of believing to a specific result. 
\item We consider a card deck, where two variables are defined: the value of the card $X$ (with numeric values or figure types) and the seed of the card $Y$. If the subject knows that the card is a \textit{king}, we have a cognitive process that leaves unchanged the $Y$ variable and leads to a sharp value for $X$ (attribute 'king'). 
\end{itemize}

In constructor-theoretic terms, a \textit{reversible computation} is the task of performing, with or without side-effects, a permutation $\Pi$ over some set $S$ of at least two possible attributes of some substrate
$$
\cup_{x\in S}\{x \rightarrow \Pi(x) \}
$$
For example, the swapping of two quantum states is a reversible operation. Because side-effects are allowed, this definition does not require the physical processes instantiating the computation to be reversible. A computation variable is a variable for which a reversible computation is always possible. A computation medium is a substrate with at least one computation variable.

In cognitive context, a reversible computation is a cognitive process which allows to change idea. For example, in a horse race, a subject can consider a  specific horse as the winner, and this process may change in time, considering other horses. The subject may change idea any time, until the race ends and the he knows which horse is the winner. Thus we can say that in cognition theory, a subject can generally  perform a cognitive process of permutation, if the external environment does not forbid it with a specific constraint. That's what happens when we think to something from a general point of view, without applying external constraints and considering all the possible situations. 
In cognitive theory, we have a \textit{general concept} when we consider representations of that concept which may evolve in a reversible way. Similarly, a \textit{general thought} is a belief state with at least one general concept. 

\section{Principle of locality: tasks of different subjects}
Einstein’s principle of locality guarantees that individual physical systems have states and attributes in the sense we have described, and it has a precise expression in constructor-theoretic form \cite{deutsch2015constructor}: there exists a mode of description such that the state of the combined system $S_1 \otimes S_2$ of any two substrates $S_1$ and $S_2$ is the pair $(x, y)$ of the states $x$ of $S_1$ and $y$ of $S_2$, and any construction undergone by $S_1$ and not $S_2$ can change only $x$ and not $y$. 

In cognition thoery, we can translate it in the following way: \textit{given two subjects $\textbf{S}_1$ and $\textbf{S}_2$ and the corresponding belief states $\psi_1$, $\psi_2$, any cognitive process undergone by the first subject and not by the second can change only $\psi_1$ and not $\psi_2$}. Such locality principle in cognition theory can be thus interpreted as a \textit{no telepathy} condition.
However, the prinicple may apply also to different substrates of the same subject. In this sense, the principle states that 

\section{Cloning task: rationality}
We introduce a type of task which has a fundamental role in the theory, the cloning task. It allows to capture the situations where subject's cognitive processes show specific features of coherence which are guaranteed by the cloning itself. Given a set $S$ of possible attributes $\{x\}$ of a substrate $\textbf{S}$, a \textit{cloning task} $R_S(x_0)$ is defined as a task on $\textbf{S} \bigoplus \textbf{S}$ which is able to take two sets of attributes $\{x\}$ and $\{x_0\}$, where $x_0$ is  some fixed (independent of $x$) attribute with which it is possible to prepare $\textbf{S}$ from generic, naturally occurring resources, and transform them in two sets  $\{x\}$:
\begin{equation}\label{cloning_attributes}
R_S(x_0)=\cup_{x\in S}\{(x,x_0) \rightarrow (x,x)\}
\end{equation}
A set $S$ of attributes is \textit{clonable} if a cloning task $R$ exists of it. Similarly we can define clonable variables and substrates.

In cognitive theory, a cloning task is a cognitive process which is able to duplicate a set $\{\psi, \phi, \xi, ...\}$ of belief states.
\begin{equation}\label{cloning_states}
\{\psi, \phi, \xi, ...\} \rightarrow  \{\psi,\psi,\phi,\phi,\xi, \xi, ... \}
\end{equation}
What makes a set of belief states clonable? It is not easy to give an answer to such a question, while we can more easily describe the features that a clonable set of belief states must show: for example, the possibility to communicate the belief state. However, the fact that we can speak about something does not guarantee that it is a clonable concept. We can speak about simple and clearly clonable facts like \textit{the tossed coin is head}, as well as about vague concepts like \textit{this idea, happiness, love, me and you}. It is clear that we need a deeper analysis to understand if the clonability property is true or not in a specific situation.  We expect that complex and unsharp concepts may evidence non-clonability. From an intuitive point of view, we can imagine that there are situations where a thought cannot be correctly replicated and communicated: it is unique, and every attempt to reproduce it is impossible. This is true for example in visual art (the uniqueness of work of arts is a proof of uniqueness of artists thoughts: a copy of a work of Caravaggio is not a work of  Caravaggio), of for the well known psychological state of flow. 
 
A sufficient condition for clonability is the following: \textit{when subjects can manage correctly and coherently a belief state, performing an unlimited number of judgements and choices based on it, such belief state is clonable}. In fact, if the copies do not represent correctly the original state, they provide information which are not consistent with the original information. What means the possibility to manage coherently such collection of belief states will be clarified by the subsidiary theory, the cognition theory in this case. In sections \ref{sec_conj} and \ref{sec_conj_super} we will provide examples of situation where such coherence is violated: we will in fact consider a well-known cognitive fallacy, the conjunction fallacy, as an example where subjects can't coherently manage information as the result on imperfect cloning.

Constructor theory of information defines  an \textit{information medium} as a clonable substrate. Similarly we have information attribute and information variable. Thus the cognitive analogue definitions are \textit{information (or rational) beliefs}, \textit{information (or rational) features} and \textit{information (or rational) concepts}. A substrate instantiates \textit{classical information} if some information variable is sharp, and if giving it any of the other attributes in such substrate was possible. In cognitive context, we refer to a classic information as a subjective representation of a concept which is \textit{known} and \textit{allows rational cognitive processes}: in other words, the situation which more clearly allows simple and rational reasoning about known elements.

\section{Distinguishability: non-ambiguity}
Constructor theory introduces another important element, the distinguishability. Its definition avoids a recursive use of distinguishability itself, while simply focusing on the concept of clonability. A variable $X$  of a substrate $S$ is \textit{distinguishable} if there exists a task 
\begin{equation}\label{task_distinguishable}
\cup_{x\in S}\{x \rightarrow i_x\}
\end{equation}
where the $i_x$ constitute an information variable. If a pair of attributes $\{x,y\}$ is distinguishable we shall write $x \perp y$: we can also say that  $\{x,y\}$ is a distinguishable variable. In the example of a semaphore, the colors can be put in correspondence to the positions of the lights: red in the top position, amber in the middle, green at the bottom.

In cognition, distinguishability represents the possibility to think to different elements without ambiguity. In other words, it is the possibility for a subject to change a set of belief states into to another set of states which have a clear identity and thus it is clonable. In the following we provide some examples.
\begin{itemize}
\item The belief states describing the knowledge of the coin outcomes are evidently information variables, and we can easily associate them to other sets: \textit{first side} $\rightarrow$ \textit{head}, \textit{second side} $\rightarrow$ \textit{tail}. 
\item Also unsharp belief states can be distinguishable. For example belief states relevant to a specific coin outcome like \textit{optimistic} and \textit{pessimistic} can be put in correspondence to other information variables, like for example \textit{win} and \textit{loose}. 
\item The states relevant to the knowledge of concepts like \textit{cold/hot} are translated in states describing the act of touching cold/hot objects.
\item The concept of distinguishability  is strictly connected to that of metaphores: as noted in \cite{lakoff2008metaphors}, many reasons suggest that \textit{our conceptual system is largely metaphorical}. As an example, when distinguishing between a bad or a good claim, we can say that it is  \textit{indefensible} or not. It is clear that the context has nothing to do with a war, but the metaphore allows to distinguish between two kinds of claims. Imagine a culture where an argument is viewed as a dance, the participants are seen as performers: clearly, a distinguishing task could be used, involving for example \textit{undanceable} or \textit{danceable}. Another example concerning \textit{ideas} evidences different kinds of methaphores: an idea can be \textit{living/died} (ideas like people), \textit{raw/warmed-up} (ideas like food), \textit{incisive} (ideas like cutting instruments).	 
\item Belief states describing unknown concepts cannot be converted into other belief states and thus they do not allow a distinguishing task.
\item A set of direction indications like \textit{left, right, overthere} contains some elements which do not allow a clear distinction. A subject simply wouldn't know where to go. Thus such set defines a non-distinguishable conceptual representation of possible directions. In simpler words, it is a set of \textit{ambiguous} indications.
\end{itemize}

\section{Measurability: partial faithfullness}
The \textit{measurement} or \textit{test} is the process leading to an experimental outcome which is intepreted as information about the original system. For example, the triggering of particle detectors following the collision of a specified beam with a given target is interpreted as the impact of a specific particle, according to a specific model and specific assumptions. In cognition, the test is the final part of the experiment, where subjects provide an outcome after receiving some information. As a trivial example, let us suppose that in the preparation phase we say \textit{there is a car, which can be blue or red}. Thus we have defined the possible outcomes of the test. The preparation can let subjects be completely uncertain about the car's color (\textit{we don't know}), or it can add a selective information, like for example \textit{we know that the car is red}. In the test phase, we ask to the subjects the color of the car. In the first case, they will answer \textit{I don't know} (when it is an allowed answer), of when forced to give a precise answer they will give one of the two colors with 50\% of probability to be correct. In the other case, they will all answer \textit{red}, if they are faithful and they have correctly understood the task. The \textit{measurability} property allows capturing such situations where the subject can produce a sharp answer every time the knowledge state is sharp. In other words, the measurability property describes the capability of subjects to answer correctly to questions whenever they know the answer.
 
The concept of measuring task has a quite articulated definition, but it  is strictly related to that of distinguishing. We say that a set of states is \textit{measurable}  if such states continue to exist after the task and the process stores its result in a second, output belief state (which must therefore be an information medium): 
\begin{equation}\label{task_measurable}
\cup_{x \in X} \{(x,x_0)\rightarrow (y_x, 'x') \}
\end{equation}
The output substrate is initially prepared with a 'receptive' attribute $x_0$. When $X$ is sharp, the output substrate ends up with an information attribute $x$ of an output variable, which represents the abstract outcome '\textit{it was x}'. Thus, measurement is like cloning a variable  except that the output substrate is an information medium rather than a second instance of the cloned substrate. A constructor is a \textit{measurer} of $X$ if there is some choice of its output variable, labelling and receptive state, under which it is capable of performing the task of formula \ref{task_measurable}.

In cognitive context, the measurement task represents the tool for inspecting the subject's belief state. From a cognitive point of view, this means that the input belief state isn't directly accessible: we can't know what really a subject thinks about for instance the outcomes of a coin toss. But we can ask the subject to think and communicate an answer to the question '\textit{which value has X?}'. It is important to note that every measurement introduces a new substrate, that is a new element in the subject's cognitive system. This is consistent with the picture given in \cite{franco2016newtheory} when trying to design a new quantum-like cognitive theory which is able to resolve the problems of the previous models.

If the measurement task is possible, we can consider the answer as true everytime the initial belief state is sharp. In case of unsharp initial state, that is when the subject doesn't know which value of $X$ is true, the constructor theory cannot say nothing. It is the subsidiary theory that adds the rules to work in such cases. Let us now consider the measurer: it is a subject which is able to find a way to put a sharp knowledge of an event in correspondence to the sharp knowledge of another replicable event, the answer. We stress the fact that the definition is centred on sharp attributes.   A measurer of $X$ is automatically a measurer of a range of other variables, because one can interpret it as such by re-labelling its outputs. For example, a measurer of $X$ measures any subset of $X$, or any coarsening of $X$ (a variable whose members are unions of attributes in $X$).
 
We present here some examples of measurements.
\begin{itemize}
\item Subjects' answers are measurements every time the set of possible answers represents a disjoint set of elements corresponding to the knowledge of specific events. That's why we say \textit{you are not asnwering to me} every time we get a vague answer.
\item Simple conventional signs used by subjects, like \textit{ok/ko}, \textit{on/off} allow to put the knowledge of specific events in correspondence to such conventions. In other words, these conventional simple answers allow the measurer to understand the subject's knowledge.  
\end{itemize} 

A final remark about the unsharp case: in quantum theory, different measurements on identically prepared unsharp states may give different outcomes.  In fact, the experimenter can compute the frequency of the different outcomes to approximate the corresponding probabilities. In other words, different instances of the same substrate are considered.
In cognition theory, subjects do not generally produce answers just like quantum particles. Subjects generally produce a sharp output only when they are sure about the event they are considering. In case they are not sure, they implicitly compute a judgement and, if needed, their answer is relevant to the most likely outcome. However, we can imagine that subject's judgements involve implicitely mental measurements over an unsharp state in order to perform the judgement.

\section{Conjectured principles of the theory: interoperability}
We have already presented the two basic principles of contructor theory. However, in \cite{deutsch2015constructor} it is noted that, if we want to describe the properties of a specicif theory (physics for example) it is necessary to seek the specific additional constructor-theoretic principles of such theory. In \cite{deutsch2015constructor}, some additional principles have been identified, which allow to represent into a physical theory the most important properties information.
In the following we will present their analogue in cognitive context.

The first additional principle, which will be labelled as number III, is called the \textit{interoperability principle} and it is very important, even if it has not been clearly stated in the prevailing conception of fundamental physics. In cognition, such principle simply states that \textit{the combination of two belief states with information valiables $S_1$ and $S_2$ is a belief state with information valiable $S_1 \times S_2$}, where multiplication symbol denotes the Cartesian product of sets.
This principle means that belief states can be combined, forming a new more complex belief state. For example, the knowledge of the seed and of the value of cards in a deck becomes a more general cognitive state which comprehends both information variables. Without this principle, each belief state couldn't be combined with others, leaving subjects' beliefs into a fragmentary condition.

Principle IV states that \textit{if every pair of attributes in a variable X is distinguishable, then so is X}. This means that if every feature forming a  concept is non-ambiguous, the concept itself is non-ambiguous. This fact is very important in cognition, since it allows to treat aggregations of features in a choerent way. For example, let us consider the variable \textit{figures in a card deck}: since every element in such variable is distinguishable, so is the concept \textit{figure}. Thus we can consider the concepts \textit{values/figures} as two non-ambiguous elements.

Similarly principle V states that \textit{if every state with attribute y is distinguishable from an attribute x, then so is y}. This means that if all the belief states involving a specific knowledge ($y$) are distinguishable from another attribute $x$ , then $x$ and $y$ are distinguishable.

Principle VI states that \textit{any number of instances of any information medium, with any one of its information instantiating attributes, is preparable from naturally occurring substrates}. In cognitive context, the priniciple means that it is possible to prepare, from naturally occurring cognitive resources, any number of any rational beliefs.

Principle VII, in a similar way, considers potentially unlimited resources not only for building an information medium, but also for information processing. Thus it states that \textit{every regular network of possible tasks is a possible task}. In cognition this implies that cognitive processes can be combined to form a new cognitive process. This is very important, since it allows to consider specific patterns of processes.

\section{Preparation: when subjects learn}
Following \cite{peres2006quantum}, we can say that the \textit{preparation} is an experimental procedure that specifies the  state, like a recipe in a good cookbook. In quantum physics, any experiment is always performed over a set of identically prepared states. The \textit{test or measurement} is the process leading to the experimental outcome. In constructor theory, a \textit{preparation task }describes a transformation where the input information '\textit{it is x}' becomes a state of a new substrate with attribute $x$. 
\begin{equation}\label{task_pfeparation}
\cup_{'x' \in 'X'} \{('x',x_0)\rightarrow ('y_x', x) \}
\end{equation}
We can interpret such formula as follows: the input state is an abstrat variable '\textit{description of X}', while the output substrate, which is subject's belief state, is initially prepared with a 'receptive' attribute $x_0$. When '$X$' is sharp, the output substrate ends up with an information attribute $x$ of $X$, which represents the knowledge about such variable. In other words, the input belief state is the given information, while the output is the subject's belief state relevant to the same object. Thus, the preparation is like cloning a variable  except that the input substrate is an information medium rather than an instance of the main substrate. The preparation task is possible in a cognitive context if the subject allows task of formula \ref{task_pfeparation}, which means that he \textit{believes} to the input information. 
\begin{itemize}
\item If the input attribute is '\textit{the coin is head}', then the subject, if admits the preparation task, prepares a belief state describing the coin with attribute \textit{head}.
\item Let us suppose that the subject does not know which card has been drawn from the deck. If someone tells to the subject that \textit{the card taken from the deck is 5 of hearts}, then the variable \textit{value} $X$ has attribute $x=5$ and variable \textit{seed}  has attribute $y=$\textit{hearts}.
\item It the input state is unsharp like for example \textit{we don't know the coin outcome}, the preparation task cannot be performed. This means that an input state like \textit{the coin outcome is not known} does not lead to a preparation of the attributes relevant to the coin outcome. However, some additional variables representing the type of uncertainty may be prepared.   
\end{itemize}

We conclude this section remarking the importance of measurement and preparation, as two complementarty processes with an important point in common: they both involve \textit{abstraction} in that \textit{one entity is represented symbolically by another}. Formulas \ref{task_measurable} and \ref{task_pfeparation} in fact evidence that the answer and the input information are something different from the subject's belief state: the subject may decide not to believe to the input information, or to give an answer which is not coherent with his beliefs. The same happens in quantum physics: any measurement is an  \textit{occurrence of a macroscopic event}, obtained following \textit{specified preparation procedures}, which provides information in an indirect way about elusive microscopic objects such as electrons, photons, etc.  Quantum theoretical predictions rely on conceptual models about such microscopic objects. In a similar way, cognition theory makes predictions relevant to subject's answers following specific preparation procedures consisting in input information. Such predictions rely on another elusive object, the subject's thought. Nobody has never seen really a thought or an electron.

\section{Unsharp states}
Let us consider a variable $X$ for which there is an unsharp state: in such case the output variable of a measurement needs not be sharp. We first define a convenient tool, the \textit{bar operation}: let $x$ be any attribute; then $\overline{x}$ ('\textit{x}-bar') is the union of all attributes (i.e. the set of all states) that are distinguishable from $x$. For example if $x$ is the belief state describing the knowledge of a card (5 of hearts), $\overline{x}$ represents the set of belief states relevant to cards which are not 5 of hearts. 

If $X$ is a variable, principle IV allows us to assign the natural meaning $\overline{X}\equiv \overline{\cup_{x\in X} x}$. When $\overline{X}$ is empty, we call $X$ a maximal variable. Any variable of the form $\{x, \overline{x} \}$ we shall call a boolean variable. Principle IV and the deﬁnition of bar trivially imply that every boolean variable is distinguishable. In addition, every boolean variable is maximal.
If we consider a standard card deck, we have two variables $U,V$, the seed $U$ and the card value $V$. The variable $U$ defines the collection of belief states where subjects know the seed of the card: $\overline{U}$ is empty, since $U$ is a maximal variable (there are no cards with something alternative to seeds). Similarly, for the variable value.  The set of belief states for which the value is a \textit{figure} instead is not maximal, its topped-bar set defines the set of states of cards with numbers.

Now, consider an attribute $\{a\}$ in which $X$ is non-sharp: then, the output variable of every measurer of $X$ is either non-sharp, or sharp with some value $x$, where $x\in X$. That means that the measurer could mistake the attribute $\{a\}$ for one having an attribute in $X$. 
In the example of the card deck, let us consider the unsharp belief state relevant to situations where the subject only knows that the card is not a figure (it is a number). Thus, if $F$ is the \textit{figure} variable, we have that the state is in the set $\overline{F}$. But if the subject gives an answer (5 for example), this could be true or false. However, it is easy to define a boolesn variable \textit{figures/numbers} which is sharp.

\section{Observables: faithfullness}
In constructor theory, an information variable $X$ is an \textit{observable} if, whenever a measurer of $X$ delivers a sharp output ‘$x$’, the input substrate really has the attribute $x$.  A necessary and sufficient condition for a variable to be an observable is  that $\overline{\overline{x}}=x$ for all its attributes $x$. 

In cognition, a concept $X$ is an observable if the subject's answer relevant to $X$ entails that the knowledge of $X$ is really consistent with the answer. In other words, there is a correspondence between the real knowledge and the answer. Measurability means that a sharp knowledge leads to a sharp output; but if the initial knowledge is not sharp, the output produced by the measurement is not necessarily true. Observability means that a sharp output automatically determines a corresponding sharp knowledge of the initial belief state, which is the essence of \textit{faithfullness}. Observability is thus possible only if the set of possible cognitive states is suitably defined in order to avoid situations where a subject must provide a sharp output even if the input is unsharp.

\section{Types of test in cognition}\label{types_test}

Cognitive psychology experiments can perform different types of test. Given an uncertain event $X$ with $N$ possible outcomes, subjects may be asked to make judgements about it in various forms: for example, a  probability (or typicality) judgement for each outcome (or on subsets of specific outcomes), a choice of the most likely outcome or a ranking of outcomes (in ascending/descending order). Even if the choice of the most likely outcome is the simplest type, it is clear that a basic task is the probability judgement over the outcomes. In fact, subjects have to analyse the possible outcomes in order to choose the most likely or make ranking. Of course, such analysis may be an intuitive process (focusing only on a limited number of features and/or  exemplars), not an analytic and slow process. Thus an important challenge is to find a mechanism able to represent ranking tasks in terms of parallel and approximative judgements.

It is clear that a quantum model studying subjects' cognitive states will mainly focus on the judged probability, which is the basic object of study. However, it is important to consider carefully its relation with the subjects' frequency of choice, that is the decision. For this reason, given the uncertain outcome $x$, it is useful to consider different types of judgements depending on the response format: 
\begin{enumerate}
\item a number $p(x)$ taken from a range of probabilities (for example 0 to 10\%, 10\% to 20\% and so on). This number $p(x)$ is given by each subject in the test. By summing each judgement and dividing by the number of subjects we obtain $\widehat{p}(x)$, that is the \textit{mean judged probability} (relevant to outcome $x$). The variance $\Delta_{p}$ gives a quantitative information about the difference between subject's answers. For example a mean judged probability of 0.9 with zero variance means that all the subjects gave the same judged probability.
\item a binary form (for example high/low probability). This form of judgement is of course very rough, but it is the easiest form to make a judgement. When there are also only two possible outcomes ($N=2$), we have a very particular case: the judgement of high probability for an event is equivalent to a judgement of low probability for its negation. We will write the frequencies of hight/low judgements $P(H), P(L)$ respectively, where the uppercase letter indicates that this is not a subjective probability, but a frequency of subjects' answers relevant to an uncertain event. It is clear that $P(H)$ near to 1 indicates that the most part of subjects consider the event likely; thus the variance is near to 0, but the mean judged probability could be very different, when considering finer ranges of probability. 
\end{enumerate}
 This last case evidences that a choice can be considered as an indirect information relevant to a judgement. Moreover, there isn't in general a simple functional form connecting the probability judgement and the response frequency.
\section{Judgements}\label{sec_judgements}

In physics, tests allow computing event probabilities. Of course, what we obtain is the measured relative frequency of an event, but if we repeat the same experiment a large (but finite) number of times, we can expect a small difference between the measured relative frequency and the true probability. For example, we can determine the approximate probability distribution of a quantum particle $p(x)$ by repeating a position test on many identically prepared particles. More particles are involved, more accurate will be the probability estimate. Each particle is detected with a random position whose statistical distribution is given by $p(x)$.

In cognitive psychology experiments, subjects don't give a random answer just like quantum particles. Usually, cognitive tests ask subjects to give the answer they think is correct, not to give the first answer that comes into mind. As noted in \ref{types_test}, most part of exeriments in cognitive psychology involves judgements. This means that cognitive tests are generally less direct than tests in quantum physics, and they are usually designed to extract more information from subjects, even if it is the result of a cognitive elaboration. 

Let us consider an observable $X$ of a substrate $\textbf{S}$, whose attributes $\{x\}$ can be labelled by a set of integers. Of course, if the subject's  cognitive state is sharp in $X$, it can be identified by a single attribute, meaning that the subject is really knowing that the value is $x$. However, when the state is not sharp, we have to consider a new task which can help to extract the partial knowledge eventually encoded into such state. This task can be used coherently only under the hypothesis of working with an observable.

The main idea of such strategy, as described in \cite{marletto2016constructor}, is to consider multiple instances of the observable $X$.  In fact, since an observable is also an information variable, we can consider multiple instances of it. In cognitive theory, this means that the concept captured by the observable $X$ can be replicated. We thus denote by $\textbf{S}^{(N)}$ the substrate consisting of $N$ instances of the original substrate $\textbf{S}$. We can also consider the instances of the same observable as a new general observable $X^{(N)} = \{\underline{s} : \underline{s} \in  X^{(N)}\}$, where  $\underline{s} =(s_1, s_2, s_3,..,s_N)$ is a set of strings of length $N$ whose digits can take values in $X$.

In each copy the measurement performed on the unsharp state may give a different output '$x$'. Each of these outputs represent a potential value of the attribute in the unsharp state. For any attribute $x$ in $X$, we can define the task $\overrightarrow{D}(N)_x$, which counts the number of replicas that hold a sharp value $x$ of $X$:
\begin{equation}\label{task_fraction}
\cup_{\underline{s} \in X^{(N)} }  \{\underline{s} \rightarrow  f(x; \underline{s})\}
\end{equation}
where the numbers  $f(x; \underline{s})$ label the attributes of the output information variable denoting the set of fractions with denominator $N$: $f_i^{(N)} = 1/N \sum_{s_i \in \underline{s}} \delta_{x,s_i}$
In other words, $\overrightarrow{D}(N)_x$ outputs the fraction of instances of $S$ on which $X$ is sharp with value $x$. Such counting task consists in measuring the observable '$X$' on each of the $N$ substrates in $S^{(N)}$, and then adding one unit to the output substrate, initially at 0, for each '$x$' detected. 

If $N \rightarrow \infty$, the attributes output of $\overrightarrow{D}(N)_x$ define the \textit{X-indistinguishabilitiy} classes. An \textit{X-indistinguishability equivalence class} is defined as the set of all attributes with the same \textit{X}-partition of unity: any two attributes within that class cannot be distinguished by measuring only the observable $X$ on each individual substrate, even in the limit of an infinite ensemble.

In cognitive terms, such task can represent the judgement, that is a new variable $J$ whose values $\{j\}$ are attributes which describe, in a specific scale, the judged probability. A probability judgement is relevant to another external attribute. For example, if $X$ is the coin outcome, we have the attributes $\{j_0,j_1,...,j_n\}$ which describe the judged probabilities that the tossed coin is head. It is important to note that judged probabilities can be considered as mere labels and we do not require that they obey to the standard probability laws. A \textit{judging task} $J_S(x_k, j_0)$ is defined as a task which is able to take two sets of attributes $\{x\}$ and $\{j_0\}$, where $j_0$ is  some fixed (independent of $j$) attribute with which it is possible to prepare $S$ from generic, naturally occurring resources and $x_k$ is the judged attribute in $\{x\}$, and transform them in two sets  $\{x, j(k)\}$, where $j(k)$ the judged probability :
$$
J_S(x_k,x_0)=\cup_{x\in S}\{(x,x_0) \rightarrow (x,j(k))\}
$$

\section{Superinformation}\label{sec_superinformation}
A superinformation medium $\textbf{M}$ is an information medium with at least two information observables that contain only mutually disjoint attributes and whose union $S$ is not an information observable. 

It is clear that  quantum information is an instance of superinformation. For example, in quantum physics, any set of two orthogonal states of a qubit constitutes an information observable, but no union of two or more such sets does: its members are not all distinguishable. In cognitive context, superinformation can be obtained by considering two information observables (thus in a situation of faithfullness) that contain only mutually disjoint attributes and whose union is not an information observable (thus it is not even a clonable variable). Our main task is to study if, also in cognition, superinformation is needed, looking for situations where the union of two concepts is not clonable. 

In the following we present some important facts relevant to superinformation, as described in \cite{deutsch2015constructor}, which could be useful also in cognitive context.
\begin{itemize}
\item  Not all information attributes of a superinformation medium are distinguishable
\item It is impossible to measure whether the observable $X$ or $Y$ is sharp, even given that one of them is.
\item Superinformation cannot be cloned
\item Pairs of observables are not simultaneously preparable or measurable 
\end{itemize}

\section{Conjunction fallacy}\label{sec_conj}
The \textit{conjunction fallacy} is a very robust effect and it is also one of the first cognitive errors studied in quantum-like cognition framework. It is particularly intriguing because it considers a basic logical operation, the conjunction of two events.

The most famous example is the Linda experiment \cite{Kahn1972}. When faced with the description of a character, Linda (who is 31 years old, single, outspoken, and very bright, with a major in philosophy; has concerns about discrimination and social justice; and was involved in anti-nuclear demonstrations while a university student), most people ranked the statement \textit{Linda is a bank teller and is active in the feminist movement} as more probable than \textit{Linda is a bank teller}, contrary to the conjunction rule. Findings about conjunction fallacy are very robust and occur with various types of stories and experimental formats. 

We can identify two main types of conjunction fallacy experiments, which may help to understand different cognitive mechanisms leading to this fallacy. The Linda experiment follows the \textit{M–A paradigm}, which connects a \textit{model} $M$ (i.e., Linda's description plus the bankteller hypothesis) with an \textit{added conjunct} $A$ (being a feminist activist). The \textit{A–B paradigm} instead does not provide specific information conveyed at the outset to describe or evoke a model. This paradigm presents a first hypothesis $A$ providing a plausible cause or motive for $B$ (i.e., the 'basic' hypothesis of interest, which is displayed both in isolation and within the conjunctive statement). As an example, in \cite{Kahn1972} the hypothesis that a randomly selected adult male \textit{has had
one or more heart attacks and is older than 55} is judged as more probable
than the simpler hypothesis \textit{has had one or more heart attacks} (the so-called health
survey scenario). Another example of \textit{A-B} paradigm is \textit{Scandinavia problem} \cite{tentori2013determinants} where the question is \textit{suppose we choose at random an individual from the Scandinavian population. Which do you think is the most probable?} and the possible choices are \textit{the individual has blond hair} or  \textit{the individual has blond hair and blue eyes}  or \textit{the individual has blond hair and does not have blue eyes}. The most part of subjects choose the conjunction, even if the concept of 'probable' is replaced by a frequency format. 

These fallacies seem to be not simply the result of misunderstanding the meaning of probability, because they even occur with bets in which the word probability never appears. Moreover, recent studies have confirmed previous partial results that also double conjunction fallacies may occur. These describe situations where each hypothesis is singularly considered less likely than their conjunction. 

Experiments of conjunction fallacy consider the following basic definitions. The \textit{mean conjunction error} is the difference between the main judged probability of the conjunction of the events and the judged probability of an event. Thus it captures the mean error evidenced by any single subject. The \textit{percentage of conjunction errors} is the percentage of subjects for which the conjunction error was not null. Of course, there may be situations where the mean conjunction error is low but the percentage is high, as well as situations where the mean conjunction error is high but the percentage of conjunction errors is not so high. When we use typicality judgements, we speak of \textit{mean conjunction effect} and similarly \textit{percentage of conjunction effects}. Another measurement of the conjunction fallacy is based on ranking probabilities relevant to a single event or to the conjunction of two events. Instead of establishing the judged probabilities, subjects have simply to rate them, deciding the most probable event. The percentage of subjects which consider the conjunction as more probable can be defined as \textit{percentage of conjunction ranking errors}.

The experiments focused on the conjunction fallacy have considered combinations of \textit{unlikely}/\textit{likely} events (\textit{U} stands for the unlikely event, while \textit{L} for the likely one). It results that the mean conjunction errors  are greater in situations \textit{UL} than in \textit{LL}. Coherently, the percentage of conjunction errors is reduced when likely events were conjoined, that is the \textit{LL} case. Moreover, it seems that the mean judged probability of a conjunction of events is mainly determined by the least likely event.

For what concerns double conjunction errors, first experimental data showed that double errors appear to be confined to \textit{LL} conjunctions, being extremely rare in \textit{LU} and \textit{UU} contexts \cite{fisk1996component}. 

Actually, predictive models for the conjunction fallacy are very limited and lack of generality. From one side, quantum-like actual models for conjunction fallacy are based on non-commuting operators acting on the same Hilbert space. This allows to produce some predictions relevant to the single conjunction fallacy, even if the free parameter corresponding to the interference factor remains not fully explained. More importantly, such model in not able to capture the double conjunction fallacy, opening the possibility to build more structured quantum-like models as suggested in \cite{franco2016newtheory}. On the other side, however, the approach based on inductive confirmation seems to suggest a general approach and a functional form of the conjunction errors, even if without a clear and formal derivation. 

\section{A constructor-theoretic model for conjunction fallacy}\label{sec_conj_super}
Using the previous definitions, we describe now a physical model for conjunction fallacy in a constructor theoretic form. We will recall the main concepts of constructor theory of information, which we will use as assumptions of the model. 

The \textit{first assumption} of the model is that a subject which is performing a judgement can  be described as a constructor, whose action produces transformations over some substrates. Such substrates are physical systems inside the subject: the brain, but also any other part of the body which somehow influences the cognitive processes. We thus do not make additional hypotheses on the form and structure of substrates: we only study the form of trasformations, and their possibility to occur.

The \textit{second assumption} is that the events that are judged in conjunction fallacy experiments  can be described as attributes, that are specific configurations of substrates. In other words, every time a subject judges an event like for example "Linda is bankteller" there must be a specific configuration of a part of the brain where such event is represented. 

The \textit{third assumption} is that such events form a distinguishable variable, as defined in formula \ref{task_distinguishable}. This means that experiments require to perform judgements on sufficiently unambiguous events.

The \textit{fourth assumption} is that subjects are able to access information they may have. This means that they are not distracted and that there aren't cognitive blocks impeding subjects from performing a measurement. We have already presented the measuring task in formula \ref{task_measurable}, which represents the cognitive trasformations allowing subjects to extract information from a substrate and to store it into another. Clearly, such task produces always the same result for sharp states on the same measured variable, while in other cases it may produce different results.

The \textit{fifth assumption} is that subjects have no interest in providing a wrong answer. In other words, when they say that they know something, they really think this. In this case, we say that the cognitive variable is an observable, which is the ideal context for an experiment.

The \textit{sixth assumption} is that subjects, in specific situations, are able to perform judgements  based on statistically independent samples. This is equivalent to the assumption that a cloning task on the original substrate is possible: judgements are described in terms of a task which performs measurements on a potentially infinite number of copies of a substrate, and then counts the number of copies where the output is a specific value.
The judgement of event  \textit{X} (which can only be true or false), called $J(X)$, is obtained from the sample proportion of\textit{N} parallel and independent simulations of \textit{X} where the measurement of \textit{X} has 1 value. We underline the fact that for each copy a measurement is performed, leading to a specific value of the measured variable. A judgement $J(X)$ identifies an $X$-indistinguishability class $[f_x]_{x\in X}$: thus $J(X)$ is a more primitive object than a probability.

\subsection{The fully rational case}
When assumptions 1-6 are true for a specific variable, we call the subject a \textit{rational reasoner}. We prove that, when all such assumptions are true in judging two events and their conjunction,  subjects can't produce conjunction fallacy.

In conjunction fallacy experiments, there are two hypotheses $X$ and $Y$, which can be considered as information variables with attribute values $\{x\}$, $\{y\}$ respectively. For example, the variable $X$ could be \textit{Linda is feminist}, with two attributes \textit{true} and \textit{false}, and similarly for $Y$ (\textit{Linda is bankteller}). The hypotheses are also assumed to be observables, since we expect that subjects' certainty about an hypothesis correspond to a real knowledge about such hypothesis. Moreover, we assume that also  the combination of the two hypotheses is again an observable. In other words, $X \cup Y$  and  $X \cap Y$ are also observables.

We thus consider $N$ instances of such system. In each instance we can perform a single measurement, as well as a joint measurement leading to a value '$x \wedge y$' of the observable. But since also $X$ and $Y$ are information variables, the same measurement in each copy must provide output values '$x$' and '$y$' consistent with '$X \cap Y$'. In simpler words, subjects are able to imagine, for each instance, the output relevant to the single events as well as their conjunction.  The task defined in formula \ref{task_fraction} thus allows producing judgements both for the single hypotheses $X$, $Y$ and for their conjunction, simply computing the proportion of attributes: it is clear that for each instance the conjunction is true only if both hypotheses are true, thus the proportion of instances where the two hypotheses are true must be lower or equal than the analogue proportion for only one hypothesis, contradicting the conjunction fallacy results. This proves our statement that assumptions 1-6 for judgements of two events and also of their conjunction are in contradiction with conjunction fallacy.

When conjunction fallacy happens, one of the six assumptions is invalid: we exclude assumptions 1, 2 and 3, because they describe minimal requirements for performing experiments. Also requirement 5 is assumed to be true, because there are no reasons for subjects to lie. Thus we can restrict ourselves to assumptions 4 and 6, which describe problems in measuring or in cloning respectively.

\subsection{The noisy case}
Let us first assume that assumption 6 is true but 4 is false. This situation reproduces what is described in \cite{costello2017explaining}: subjects are rational, but they can make mistakes in measurements.
This may happen simply beacuse human brain is not an ideal system, and each sampling on a copy can sometimes produce a wrong result.

This model is also called \textit{probability with noise model (PN)} and it has been used to propose a rational  description of conjunction fallacy. However, when the participants are asked to judge events and their combination,  all of the samples used to estimate the proportions for single events should be pooled with all of the samples used to estimate the conjunction into one common collection of samples. 

Such random noise cannot in general produce conjunction fallacy. In fact, random noise  acts both for single and for conjunct judgements and there is no a-priori reason to make hypoteses about different strength of the random noise. 
However, when $J(X)$ is near to 0 or to 1, the effect of noise on judgement is biased in one direction and, if its entity is sufficiently strong, it may produce conjunction fallacy. However, this situation is very specific, since it is relevant only to judgements of extremely likely/unlikely events, while conjunction fallacy is observed in a wider range of situations.

\subsection{The no-cloning case}
We now consider the case where only assumption 6 is wrong when considering the combination of concepts, which means that such combination is not an observable, but a superinformation variable.
Of course, in real cases we should also conisder the presence of random noise in measurements.

The fact that the combination of observables $X$ and $Y$ is not an observable clearly means that subjects in this case cannot manage information in a fully rational way, since they can't work with copies of the original substrate. This also means, as descibed in section \ref{sec_superinformation}, that such pair of observables is not simultaneously preparable or measurable. 

We have thus shown that the attempt of describing in a general way the conjunction fallacy leads to consider cognition theory as an instance of a superinformation theory. Also quantum theory of information is an instance of superinformsation theory. Thus it results natural to look at the mathematical framework of quantum theory as a possible toolbox for models in cognition theory, even if at the moment this is only a possibility. For example, in an extremal case, certainty about a first hypothesis $X$ may lead to a complete uncertainty about the second hypothesis $Y$ (incompatible variables): this case represents a situation of extremely limited resources, which we have already put in correspondence to rational ignorance strategy \cite{franco2007quantum}.  

On the opposite side, we can consider situations where subjects are able to find sharp measurements both for $X$ and $Y$. For example, given Linda's description, we could accurately consider combinations of events where a person is certainly feminist and bankteller, without producing the conjunction fallacy. This reasoning strategy can be called \textit{fully rational strategy}, which requires a big amout of cognitive resources to manage all the information. In this regime, the union of observables is again an observable. On the contrary, the bounded rationality strategy \cite{simon1990bounded} represents situations where intuitive judgements make use of superinformation.   

In conclusion, we have provided a physical general model in constructor theoretic terms for joint judgements and we have shown that superinformation in a necessary element in order to reproduce conjunction fallacy.  

\section{Conclusions}
This article provides solid arguments showing that human cognition needs to use conceptual instruments whose nature is more general than those coming from classical information. This fact is a consequence of precise results of constructor theory, declinated onto cognitive psychology. 

However, the need to use superinformation media to represent wide classes of situations does not directly justify us to use a quantum-like theory, that is a  probability framework for cognition. We only know that cognition theory and quantum theory are subsidiary theories of the same class of superinformation theories. They work on different kinds of substrates and they may need different formalisms. 

This paper represents a first step to the use of constructor theory for cognition. In fact, a subsidiary theory of cognition needs to be build, following the constraints given by experimental results. We expect that various known cognitive heuristics, like for example gambler's fallacy \cite{gilovich2002heuristics} or the hot-hand fallacy \cite{Burns1} will confirm the superinformation nature, suggesting a specific form of the theory.

 \section{Aknowledgements}
I would like to thank Mario Rasetti for his always inspiring suggestions, Jerome Busemeyer for useful discussions about cognitive processes and their quantum-like description, Vincenzo Crupi for discussions about the double conjunction fallacy and Giuseppe Pelizza for suggestions about non-clonability intepretation. Finally, a special thank goes to my wife Marina for encouraging me in following my interests and dreams, conciliating them with familiar life and work.
\footnotesize
%
%

\section{References}
\bibliography{../mybib}{}

\begin{thebibliography}{10}

\bibitem{gilovich2002heuristics}
Thomas Gilovich, Dale Griffin, and Daniel Kahneman.
\newblock {\em Heuristics and biases: The psychology of intuitive judgment}.
\newblock Cambridge university press, 2002.

\bibitem{franco2009conjunction}
Riccardo Franco.
\newblock The conjunction fallacy and interference effects.
\newblock {\em Journal of Mathematical Psychology}, 53(5):415--422, 2009.

\bibitem{busemeyer2011quantum}
Jerome~R Busemeyer, Emmanuel~M Pothos, Riccardo Franco, and Jennifer~S
  Trueblood.
\newblock A quantum theoretical explanation for probability judgment errors.
\newblock {\em Psychological review}, 118(2):193, 2011.

\bibitem{pothos2009quantum}
Emmanuel~M Pothos and Jerome~R Busemeyer.
\newblock A quantum probability explanation for violations of �rational�
  decision theory.
\newblock {\em Proceedings of the Royal Society of London B: Biological
  Sciences}, pages rspb--2009, 2009.

\bibitem{pothos2011quantum}
Emmanuel~M Pothos and Jerome~R Busemeyer.
\newblock A quantum probability explanation for violations of symmetry in
  similarity judgments.
\newblock In {\em Proceedings of the 32nd Annual Conference of the Cognitive
  Science Society}, page 284854, 2011.

\bibitem{wang2014context}
Zheng Wang, Tyler Solloway, Richard~M Shiffrin, and Jerome~R Busemeyer.
\newblock Context effects produced by question orders reveal quantum nature of
  human judgments.
\newblock {\em Proceedings of the National Academy of Sciences},
  111(26):9431--9436, 2014.

\bibitem{boyer2016testing}
Thomas Boyer-Kassem, S{\'e}bastien Duch{\^e}ne, and Eric Guerci.
\newblock Testing quantum-like models of judgment for question order effect.
\newblock {\em Mathematical Social Sciences}, 80:33--46, 2016.

\bibitem{khrennikov2014quantum}
Andrei Khrennikov, Irina Basieva, Ehtibar~N Dzhafarov, and Jerome~R Busemeyer.
\newblock Quantum models for psychological measurements: an unsolved problem.
\newblock {\em PloS one}, 9(10):e110909, 2014.

\bibitem{pleskac2013s}
Timothy~J Pleskac, Peter~D Kvam, and Shuli Yu.
\newblock What's the predicted outcome? explanatory and predictive properties
  of the quantum probability framework.
\newblock {\em Behavioral and Brain Sciences}, 36(03):303--304, 2013.

\bibitem{franco2016newtheory}
Riccardo Franco.
\newblock Towards a new quantum cognition model.
\newblock {\em arXiv preprint arXiv:1611.09212v1}, 2016.

\bibitem{deutsch2015constructor}
David Deutsch and Chiara Marletto.
\newblock Constructor theory of information.
\newblock 471(2174):20140540, 2015.

\bibitem{deutsch2013constructor}
David Deutsch.
\newblock Constructor theory.
\newblock {\em Synthese}, 190(18):4331--4359, 2013.

\bibitem{huth2016natural}
Alexander~G Huth, Wendy~A de~Heer, Thomas~L Griffiths, Fr{\'e}d{\'e}ric~E
  Theunissen, and Jack~L Gallant.
\newblock Natural speech reveals the semantic maps that tile human cerebral
  cortex.
\newblock {\em Nature}, 532(7600):453, 2016.

\bibitem{murphy1985role}
Gregory~L Murphy and Douglas~L Medin.
\newblock The role of theories in conceptual coherence.
\newblock {\em Psychological review}, 92(3):289, 1985.

\bibitem{broughton2004faceted}
V~Broughton.
\newblock Faceted classification.
\newblock {\em Essential Classification}, pages 257--83, 2004.

\bibitem{rosenfeld2002information}
Louis Rosenfeld and Peter Morville.
\newblock {\em Information architecture for the world wide web}.
\newblock " O'Reilly Media, Inc.", 2002.

\bibitem{ranganathan1985faceted}
Shiyali~R Ranganathan.
\newblock Faceted analysis.
\newblock {\em En Chan, LM et al., eds. Theory of subject analysis. Littleton,
  CO: Libraries Unlimited}, pages 86--93, 1985.

\bibitem{lakoff2008metaphors}
George Lakoff and Mark Johnson.
\newblock {\em Metaphors we live by}.
\newblock University of Chicago press, 2008.

\bibitem{peres2006quantum}
Asher Peres.
\newblock {\em Quantum theory: concepts and methods}, volume~57.
\newblock Springer Science \& Business Media, 2006.

\bibitem{marletto2016constructor}
Chiara Marletto.
\newblock Constructor theory of probability.
\newblock 472(2192):20150883, 2016.

\bibitem{Kahn1972}
Daniel Kahneman and Amos Tversky.
\newblock Subjective probability: A judgment of representativeness.
\newblock {\em Cognitive psychology}, 3(3):430--454, 1972.

\bibitem{tentori2013determinants}
Katya Tentori, Vincenzo Crupi, and Selena Russo.
\newblock On the determinants of the conjunction fallacy: probability versus
  inductive confirmation.
\newblock {\em Journal of Experimental Psychology: General}, 142(1):235, 2013.

\bibitem{fisk1996component}
John~E Fisk and Nick Pidgeon.
\newblock Component probabilities and the conjunction fallacy: Resolving signed
  summation and the low component model in a contingent approach.
\newblock {\em Acta Psychologica}, 94(1):1--20, 1996.

\bibitem{costello2017explaining}
Fintan Costello and Paul Watts.
\newblock Explaining high conjunction fallacy rates: The probability theory
  plus noise account.
\newblock {\em Journal of Behavioral Decision Making}, 30(2):304--321, 2017.

\bibitem{franco2007quantum}
Riccardo Franco.
\newblock Quantum mechanics, bayes' theorem and the conjunction fallacy.
\newblock {\em arXiv preprint quant-ph/0703222}, 2007.

\bibitem{simon1990bounded}
Herbert~A Simon.
\newblock Bounded rationality.
\newblock In {\em Utility and probability}, pages 15--18. Springer, 1990.

\bibitem{Burns1}
Bruce~D Burns and Bryan Corpus.
\newblock Randomness and inductions from streaks: Gambler's fallacy versus hot
  hand.
\newblock {\em Psychonomic Bulletin \& Review}, 11(1):179--184, 2004.

\end{thebibliography}
\bibliographystyle{unsrt}
\end{document}